\documentclass[11pt,a4paper]{article}
\usepackage{latexsym}
\usepackage{hyperref}
\usepackage{url}

\usepackage{algorithm}
\usepackage{algpseudocode}
\usepackage{graphicx}
\usepackage{tabularx}
\usepackage{booktabs}

\usepackage{natbib}
\title{An exploratory study of L1-specific non-words}
\author{David Alfter}
\date{}

\begin{document}

\maketitle

%Including the abstract exceeds the page limit
\begin{abstract}
    In this paper, we explore L1-specific non-words, i.e. non-words in a target language (in this case Swedish) that are re-ranked by a different-language language model. We surmise that speakers of a certain L1 will react different to L1-specific non-words than to general non-words. We present the results from two small case studies exploring whether re-ranking non-words with different language models leads to a perceived difference in `Swedishness' (pilot study 1) and whether German and English native speakers have longer reaction times in a lexical decision task when presented with their respective L1-specific non-words (pilot study 2). Tentative results seem to indicate that L1-specific non-words are processed second-slowest, after purely Swedish-looking non-words. 
\end{abstract}

% PARTIM I - General Swedish non-words
\section{General Swedish non-words}
Non-word, i.e. strings of characters that look like words but actually are not, are often used for diagnostic testing purposes as in LexTale \citep{lemhofer2012introducing} or in psycholinguistic studies. In the past, researchers often came up with non-words that suited their specific purposes by changing single letters in existing words. However, this is a very subjective judgment of goodness and often there is a rather specific use-case beyond which the non-words have little use.

Besides this manual generation of non-words, there are some resources which allow for lookup or generation of non-words. For English, there is the ARC database which contains 358.534 monosyllabic non-words \citep{rastle2002358}. It offers a very fine-grained search engine for selecting non-words. There are also other resources such as WordGen \citep{duyck2004wordgen} and Wuggy \citep{Keuleers2010} which allow for the generation of (non-)words in Dutch, English, German, French, Spanish, Serbian, and Basque. However, Wuggy requires a syllabified corpus in order to learn how to generate words. Another drawback with most non-word generation programs is that they generate non-words but do not rank them according to goodness; this is often left to the researcher.
% Dutch, English, German, and French
Recent work by \citet{osama2018automatic} focuses on generation of English non-words but additionally investigates how automatically generated non-words can be assessed for goodness in an objective manner.

Non-words are quite language-specific, as a certain non-word in one language can be a valid, existing word in another language. To the best of our knowledge, there is no previous work on Swedish non-words.

In this work we first propose a general non-word generation pipeline for Swedish non-words that builds on and extends previous efforts. We then investigate the potential of L1-specific non-words in the target language Swedish, i.e. non-words that not only are valid letter combinations according to the Swedish language but additionally also according to another language. 
% We surmise that...
%\subsection{Research questions} % Research questions

For L1-specific non-words, we explore two research questions.
% RQ1: can Swedish-looking non-words be perceived as not Swedish?
Research question one asks whether non-words generated based on a Swedish language model but ranked by a non-Swedish language model can be perceived as less Swedish.
% RQ2: do L1-specific non-words result in longer response time?
Research question two asks whether L1-specific non-words result in longer response times in speakers of that specific L1, in comparison to general Swedish non-words.
% => real-word recognition should be fast for native7
We present the results from two pilot studies on L1-specific non-words that aim at answering these research questions.

\section{Methodology}
For non-word generation, we use a three-step approach. The first step is generation, the second step is filtering and the third step is ordering/ranking. 

We use an exhaustive generation approach for candidate generation of non-words between length two and up to length five, i.e. we generate all possible combinations (with repetition) of letters up to length five. We use a random non-exhaustive generation for candidate generation of non-words of length six to eleven, as the number of possible combinations increases from about 20 million for sequences of length five to about 600 million for sequences of length six. 

Candidates of length six and up are generated using a position-aware character-based 4-gram Swedish language model trained on newspaper and fiction corpora
% TODO insert what it is trained on
. The model is position-aware inasmuch as it learns which characters tend to typically occur at the front, middle and end of words. 

When generating non-words, the first character is chosen randomly. 
% TODO shouldnt this always be START-char ??
After that, the language model uses n-gram probabilities to predict the next character(s) according to the %pseudo-random algorithm  
algorithm presented in \ref{algo:pr}. The model uses the highest available n-gram order at each step and stops when the length of the sequence matches the required target length. 
% This can be further refined, see future work

% algorithm here:
% order all possible continuation candidates by likelihood
% go through list with 50% chance of choosing candidate at position x
% if no candidate after list exhaustion, use last candidate
\begin{algorithm}
\caption{Generating sequences}
\label{algo:pr}
\begin{algorithmic}
\Function{next}{$s$}
\State $c$ $\gets$ cont($s$)
\Comment $c$ is continuation set% of continuation characters
\State $c$ $\gets$ order($c$)
\ForAll{$x$ in $c$}
\State $r\gets$ random number
\Comment between 0 and 1
\If{$r \geq 0.5$}
\State \Return $x$
\EndIf
\EndFor
\State \Return last($c$)
\EndFunction
\end{algorithmic}
\end{algorithm}

With $s$ being the current \emph{stub}. The stub consists of the last $n$ generated characters, ranging from one character at the beginning of generation to three characters after having generated at least 3 characters. After this, the stub will always be of length 3 and the 4-gram language model will predict the continuation character.
In algorithm \ref{algo:pr}, \texttt{cont($s$)} is the continuation function that, based on the length of $s$, returns the set of all continuation characters for the stub $s$ according to the current n-gram language model. Continuation characters are single characters that have been observed as following after the stub in the training data. After that, continuation characters are ordered by their log-likelihood, from most probable to least probable. Then, for each continuation character, we return it with a probability of 50\% or go down to the next character, which again has a 50\% chance of being returned, etc. If the loop terminates without returning, the last (and least probable) continuation character is returned.
In case the continuation function returns the empty set, i.e. no continuation characters have been observed for the current stub, the method aborts, a new stub is randomly generated and generation is attempted again. 

% exhaustive generation for up to len 5
% cutoff 10k for 4, 18k for 5
% därefter: random non-exhaustive trigram stub generation + q-gram stub expansion
After the generation step 
%for words up to length five
, we filter out all words that are present in the 1 million word form database Saldo's morphology\footnote{The resource contains roughly 2 million word forms but only 1 million \emph{unique} word forms.} (Saldom) \citep{borin2013saldo}. We also exclude common names and abbreviations.
For words of up to length five, we also exclude all words that have a low language model probability, such as ``xxxxx'' or ``\"a\aa \aa \aa \aa''. 

% manual selection of 2-gram
% ?? discard 2-grams??

% filtering
% - saldom
% - low LM prob (<11)
% - names
% - abbreviations
% - phonotactics? -> inherently captured by LM, also Future work
% -- disallow ending in c

% ordering
% - lm 4-gram
% - # orthographic neighbors?? -> exclusion of no orthographic neighbors; more neighbors => more confusability
% => as in Hamesh, we have found neighborhood to be rather meh

Next, all candidates are ordered using the position-aware language model for Swedish. We experimented with orthographic neighborhood size but, as in \citet{osama2018automatic}, we found this feature to be low-performing and discarded it. The only ordering criterion is the n-gram probability.

% POS assignment!
% CEFR levels!
% Dictionary entries?
In contrast to previous work on automatic non-word generation, we assign each word a part-of-speech and a target proficiency label.

% LSTM encoder trained on lemma-pos (lexin?)
For part-of-speech assignment, we train an LSTM sequence--to--multi-label neural network that captures character-based position-specific features. In this way, we hope to learn sensible word to part-of-speech associations.
% sequence to multilabel classification
% -> learn sensible associations 

% character based POS mapping . seems to capture inflected form information (not present in training data?)
We also assign each generated non-word a target CEFR level, i.e. the level at which a learner of this level should be able to deal with the word. We expect that extending non-words with such information can lead to more sensible non-word exercises for learners of different levels
. For this, we adapt previous work by \citet{alfter2018towards} that assigns each word a target label.

% PARTIM II - From L1-agnostic to -specific
\section{L1-specific Swedish non-words}
%related work/motivation
After the generation of Swedish non-words, we want to explore whether we can use different L1 language models to re-rank the Swedish non-words for learners of Swedish with different L1 background. The intuition is that for example in lexical decision tasks, if a non-word is perceived both as having Swedish qualities as well as having qualities of one's own L1, the decision takes longer and requires a higher proficiency of Swedish. Faster correct decisions could potentially yield more expressive power than general non-word tests.

We have formulated two research questions in relation to L1-specific non-words, namely:
\begin{enumerate}
    \item Can Swedish non-words be perceived as less Swedish?\label{rq1}
    \item Do speakers of L1 $X$ react more slowly on non-words that look Swedish but also $X$?\label{rq2}
\end{enumerate}

% The following section presents related work and then two small case studies aiming at hinting at the possible answers to \ref{rq1} and \ref{rq2}

\subsection{Related work}
% Modeling Reading: The Dual-Route Approach, Max Coltheart
The theory of the dual-route approach to reading aloud seems to have been first proposed by de Saussure in 1922 \citep{saussure1983course}, formalized by \citet{forster1973lexical} and later adapted by \citet{baron1977mechanisms} to also cover reading comprehension. 
%  dual-route for reading comprehension:
%Baron (1977),  Forster and Chambers 73, de Saussure (1922; translated 1983, p. 34)
%    - lexical route => use visual information to retrieve word
%    - non-lexical route: phonological route => transform visual information into phonological form and retrieve
The theory states that when encountering a word in reading, two routes are activated simultaneously: the lexical route uses only the visual stimulus from the word to retrieve the word from the mental lexicon, if it can be found in the mental lexicon. The non-lexical route uses grapheme-to-phoneme mappings to transform the visual stimulus into a mental phonological representation which is then used to either pronounce or retrieve the word from the mental lexicon. The result of encountering a word in reading is then whichever of the two processes returns a result first.
% DRC is computational model of theory
The theory was later adapted into a computational model by \citet{coltheart2001drc}.

% Activation of lexical route for non-words => activate similar words
% For example, a nonword like SARE which is similar to  many  entries  in  the  orthographic  lexicon  will  be  read  aloud  with  a  shorter  reaction time (RT) than a nonword like ZUCE which is similar to few (McCann & Besner, 1987). (p.13)
Furthermore it as been shown that if a non-word is orthographically similar to many entries in the mental lexicon, its reading time will be faster than if is has few or no orthographic neighbors in the mental lexicon \citep{mccann1987reading}.
Since non-words activate similar words in the mental lexicon, orthographic neighborhood could be interesting in generating qualitatively high non-word lists. Indeed, if a non-word has a lot of potential words it could be confused with, it would increase cognitive load when trying to decide whether the non-word is a word or not.
% Activation of similar words in mental lexicon
% => use neighborhood size
% but: The number of orthographic neighbors a pseudohomophone has does not influence how fast it is read aloud (p.15)
However, it has been found that the number of orthographic neighbors of non-words does not influence response time in lexical decision tasks \citep{coltheart2005dual}. This is consistent with the findings in \citet{osama2018automatic} where they found neighborhood size to be performing rather modestly in ranking non-words for word-likeness.
% => consistent with O and Zesch

With bilinguals, it has been shown that in monolingual lexical decision tasks where non-words can be valid words in their other language or at least form valid letter combinations in the other language, response time was slower on these non-words, suggesting an activation and inhibitory effect of the other language system \citep{thomas2000language}.

%-------
\subsection{Pilot study 1}
% Swedish nonword list for speakers of Arabic
% idea: using X language model for speaker of X, might introduce more confusing items
In the first pilot study, we want to answer the question whether Swedish non-words re-ranked by an Arabic language model are perceived as more Arabic.

Written Arabic is typically not vocalized (i.e. short vowels are not indicated in writing). However, in order to make the two language models similar to each other, we need vocalizations in Arabic. We used the vocalized Tashkeela corpus
% approximate transliteration of vocalized corpus
\citep{zerrouki2017tashkeela}. We use a heavily simplified non-standard transliteration scheme in order to have more overlap between the two language models. 
% small scale/pilot study of influence of ranking LM
% use 6-grams => random generation using SV LM

For the study, we used non-words of length six. 
We generate 10000 random non-words of length six. We then rank this list with different language models as follows:
first, we select the top twenty words as ranked by the Swedish language model (group 1). Next, we select the top twenty non-words as ranked by the Arabic language model (group 2). Finally, we also take the intersection of the first 1000 items in both lists, which gives us 138 items. From this set, we select twenty random words for inclusion in the study (group 3).
% use SV LM for ranking
% use AR LM for ranking
% take first 20 of each
% take intersection of first 1000 of both => 138 items
% select 20 (random, because type is set())
% ask people to rate words as more SV or AR

Using a Google Forms form, we ask people to indicate for each word whether it looks more Swedish or more Arabic. The words are ordered in the following manner: first a word from group 1, then a word from group 2, then a word from group 3, then a word from group 1, and so on until the end of each list. All the words were presented as a single list on one page. The task was not timed.

The number of participants was five, with two of the participants speaking Arabic as L1 but having no knowledge of Swedish, two participants speaking Swedish as L1 with no knowledge of Arabic and one participant speaking Swedish as L1 and having advanced knowledge of Arabic as L2.

% expected outcome: SV LM => SV, AR LM => AR, intersection => mixed
The expected outcome of the study is that the non-words ranked by the Swedish model are perceived as more Swedish (group 1), non-words rated by the Arabic language model are perceived as more Arabic (group 2) while items in group 3 could be freely assigned to either language since they share orthographic features of both languages.
% => what does this tell us??

% results
Table \ref{tab:res0} shows the results. The table shows how many words of each group were rated as either Swedish or Arabic by the majority of voters.

\begin{table}
\begin{tabularx}{\textwidth}{XXX}
\toprule
& Swedish & Arabic \\
\midrule
%\multicolumn{3}{>{\hsize=3\hsize}X}{\textbf{Majority}} \\

Group 1 & 14 & 6 \\
Group 2 & 5 & 15 \\
Group 3 & 17 & 3 \\

\bottomrule
\end{tabularx}
\caption{Swedish-ness and Arabic-ness ratings}\label{tab:res0}
\end{table}

We can see that the majority of raters rated Swedish-looking non-words as Swedish with 14 out of 20, and Arabic-looking non-words as Arabic with 15 of 20. 
%Given that we have an even number of raters, if we count all ties, i.e. cases where 3 out of 6 raters agreed, we see that there seems to be a consensus for Swedish-looking non-words to be rated as Swedish-looking (18/20) and Arabic-looking non-words to be rated as Arabic (17/20). 
Concerning group 3, which has both Swedish and Arabic features, only five words were rated as more Arabic than Swedish by the majority of raters. 
% words where the outcome is not the predicted outcome
% analysis
This seems to confirm our hypothesis that reordering Swedish non-words using a different-language language model leads to a perceived difference in Swedish-ness.
% kanden => fil kand => kander?

% -------------

% additional pilot study
\subsection{Pilot study 2}
In pilot study 2, we want to investigate how reading times differ in speakers of different L1s when confronted with L1-specific non-words.
The dependent variable is reaction time while the independent variable is L1. In order to increase the internal validity of the experiment,
the experiment itself was designed as simple as possible.
% does response time correlate with L1 specific word rankings?
% -> if presented with a swedish nonword that was ranked by L1 model, is response time higher?
% --> would speak for higher cognitive load (more confuseability) for L1 specific NW list

In the study, we target two different L1 speaker groups, namely German and English L1 speakers. 

% EN LM : NLTK Brown 
We build the English language model from the Brown corpus, accessed through the NLTK Python toolkit \citep{bird2009natural}.
% DE LM : German Mixed 2011 part of cited
We build the German language model from the German Mixed 2011 part of the corpus by 
\citet{goldhahn2012building}

For this study, we re-use the 10000 item non-word list of length six mentioned in pilot study 1. We rank the list using the Swedish, German and English language models. We take the top 20 words of each ranking, replacing words that are already in one of the other lists with words from further down the current list, i.e. if we have already collected twenty non-words for Swedish and we are working on the German list, if a word from the top twenty German list is already in the top twenty Swedish list, we ignore the word and continue down the list until we have twenty non-words for the German setting.
% add saldom words of same length
As fillers, we add twenty random existing Swedish words of length six from Saldom.

% present in blocks where each block contains in randomized order one of SV,DE,EN,saldom
The words are ordered in blocks of size four where each block contains one non-word from each language model plus a filler. The words inside each block are in random order and the words are presented one after another as a self-paced lexical decision task.
Before starting the task, participants have to indicate their mother tongue as well as their self-reported proficiency in Swedish on a 3-point scale from beginner (A1/A2) to intermediate (B1/B2) to advanced (C1/C2). The options for mother tongue are ``Swedish'', ``German'', ``English'' and ``Other''.
In order to obfuscate the true aim of the task, namely the impact of L1-specific non-words, participants are asked to identify existing Swedish words from non-existing Swedish words.
%Information about proficiency in Swedish was self-reported.

% control group: SV native speakers
% -> expected pattern of fastest for OO, followed by SV, lower for EN and DE
The control group consists of three native Swedish speakers. For the control group, we expect the lexical route to outperform the non-lexical route on existing Swedish words and thus we expect fasted reaction times on existing Swedish words and slowest reaction times on non-word rated by the Swedish language model.
% present information about informants (L1, L2 proficiency...)
For Swedish-looking non-words, we expect the control group to take longest, since non-words, by default, activate the non-lexical route.

% => non-word recognition should activate SV phonological system and result in slower time
% => non-word non-SV should be rejected more quickly by SV phonological system than non-word SV
However, response times for English- and German-looking non-words should lie in-between, as the Swedish grapheme-to-phoneme mental model should be able to discard those words as non-Swedish.

For German, we have eight participants with German as native language of which five have a high proficiency in Swedish, one informant has intermediate Swedish knowledge and one information has basic Swedish knowledge. 

For English, we have four participants with English as native language, two beginner level learners, one intermediate and one advanced.

We also have three participants whose native language was neither of the tested ones. One participant has an intermediate level of Swedish and two participants have an advanced level of Swedish.

% results
\subsubsection{Results}
Table \ref{tab:res} shows the results for the control group. The table shows the average time taken (in seconds) for each rater and each of the groups: German, English, Swedish and Filler (existing words), as well as the normalized average of the three raters. The normalized average is calculated by first calculating each rater's average of averages and then dividing each rater's average by their average of averages. We then sum up averages for the different groups and divide the result by the number of raters. The normalized average is a dimensionless number.
% TODO
% This sounds rather confusing. Does it need a formula? Explanation?

\begin{table}
\begin{tabularx}{\textwidth}{lXXXX}
\toprule
Group & R1 & R2 & R3 & nAvg\\
\midrule
German  & 4.75 & 1.60 & 2.50 & 1.10 \\ %& 2.95 \\
English & 3.50 & 1.60 & 2.15 & 0.94 \\ %& 2.41 \\
Swedish & 4.85 & 1.80 & 3.05 & 1.23 \\ %& 1.71 \\
Filler  & 1.45 & 1.30 & 2.40 & 0.72 \\ %& 3.23 \\
\bottomrule
\end{tabularx}
\caption{Results: Control group}\label{tab:res}
\end{table}

As we can see from table \ref{tab:res}, the control group was always slowest on Swedish-looking non-words ranked high by the Swedish language model. Raters 1 and 2 were fastest on existing Swedish words, while rater 3 shows a slight deviance from this tendency, being slightly quicker on English words. This may well be due to faster rejection times of more non-Swedish-looking words as opposed to slower recognition times of real existing Swedish words. We will look deeper into this dichotomy for the German and English participants.
% On average however, raters were quickest on existing words and slowest on Swedish non-words. 

\begin{table}[!ht]
    \begin{tabularx}{\textwidth}{lXXXX}
    \toprule
    Rater & DE & EN & SV & FI \\
    \midrule
 R1 (B) & 72.5\% & 75\% & 47.5\% & 55\% \\
    R2 (I) & 70\% & 90\% & 85\% & 85\% \\
    R3 (I) & 85\% & 100\% & 65\% & 85\% \\
    R4 (A) & 100\% & 95\% & 100\% & 70\% \\
    R5 (A) & 80\% & 100\% & 100\% & 95\% \\
    R6 (A) & 95\% & 100\% & 95\% & 80\% \\
    R7 (A) & 90\% & 100\% & 85\% & 85\% \\
    R8 (A) & 100\% & 96\% & 100\% & 75\% \\ \midrule
    R9 (B) & 60\% & 90\% & 20\% & 80\% \\
    R10 (B) & 50\% & 65\% & 25\% & 60\% \\
    R11 (I) & 70\% & 85\% & 35\% & 90\% \\
    R12 (A) & 75\% & 80\% & 50\% & 95\% \\ \midrule
    R13 (I) & 80\% & 90\% & 45\% & 75\% \\
    R14 (A) & 60\% & 90\% & 20\% & 80\% \\
    R15 (A) & 95\% & 100\% & 95\% & 90\% \\
    \bottomrule
    \end{tabularx}
    \caption{Rater accuracy in percent}
    \label{tab:acc}
\end{table}

Table \ref{tab:acc} shows the rater accuracy by category of non-words. The headers DE, EN and SV stand for German-looking, English-looking and Swedish-looking non-words and FI stands for fillers, i.e. existing Swedish words. Accuracy indicates how many non-words raters correctly discarded as non-existing words for columns DE, EN and SV whereas it indicates how many existing Swedish words raters recognized as existing words for fillers. Raters R1 to R8 are German native speakers, raters R9 to R12 are English native speakers and raters R13 to R15 are native speakers of unspecified other languages.
As can be gathered from the table, accuracy in rejecting non-words and recognizing existing words is quite high in the German-speaking rater group. Values drop a bit for English and other-language native speakers, especially for the rejection of Swedish-looking non-words.

Table \ref{tab:res2} shows the reaction time results from the German and English native speakers, and for completeness also includes the participants whose mother tongue was neither of the targeted. Due to space constraints, the table header has been abbreviated to the format $xy$. The first letter $x$ takes the values D, E, S and F stand for German, English, Swedish (non-words) and Fillers (existing words) respectively. The second letter $y$ has either the numbers $0$ and $1$ which correspond to rejection and acceptance times or C which stands for the combined reaction time. Thus, D$0$ stands for the rejection time of German-looking non-words, F$1$ stands for the acceptance time of existing Swedish words, etc. Note that the $x$C columns are not the arithmetic mean of the $x$0 and $x$1 columns due to differing size. Values of $0$ indicate that the rater never accepted a word of this column as existing word (values of $0$ only occur in $x1$ columns). The columns `R' denotes the different raters with their respective self-indicated proficiency of Swedish (B = beginner, I = intermediate, A = advanced). We also indicate the normalized average nA (=nAvg) as well as the normalized average excluding beginner learners nA2 (=nAvg (I+A)).

\begin{table*}\noindent
\begin{minipage}[c]{\textwidth}
\begin{tabularx}{\textwidth}{XXXXXXXXXXXXX}
\toprule
R & D0 & D1 & DC & E0 & E1 & EC & S0 & S1 & SC & F0 & F1 & FC \\
\midrule

\multicolumn{13}{>{\hsize=10\hsize}X}{\textbf{German participants}} \\
\midrule
R1 (B) & 2.14 & 3.09 & 2.40 & 2.7 & 3.3 & 2.85 & 2.58 & 2.33 & 2.45 & 2.83 & 3.14 & 3.00 \\ % B7
%---
R2 (I) & 2.57 & 4.67 & 3.20 & 2.78 & 7.00 & 3.20 & 3.18 & 5.00 & 3.45 & 3.00 & 3.35 & 3.30 \\ % D4
R3 (I) & 2.29 & 5.33 & 2.75 & 2.00 & 0 & 2.00 & 4.69 & 4.00 & 4.45 & 3.67 & 2.00 & 2.25 \\ % 60
%---
R4 (A) & 4.35 & 0 & 4.35 & 4.21 & 0 & 4.21 & 4.95 & 0 & 4.95 & 4.83 & 2.64 & 3.30 \\ % G6
R5 (A) & 7.31 & 7.5 & 7.35 & 5.60 & 0 & 5.60 & 5.60 & 0 & 5.60 & 12.00 & 3.26 & 3.70 \\ % 84
R6 (A) & 2.47 & 3.00 & 2.50 & 2.65 & 0 & 2.65 & 3.53 & 7.00 & 3.70 & 2.50 & 1.65 & 1.75 \\ % 13
R7 (A) & 1.83 & 4.50 & 2.10 & 2.10 & 0 & 2.10 & 4.12 & 3.33 & 4.00 & 3.33 & 1.82 & 2.05 \\ % 24
R8 (A) & 10.20 & 0 & 10.20 & 3.89 & 7.00 & 4.05 & 8.00 & 0 & 8.00 & 14.40 & 5.33 & 7.60 \\ % C3
nA & 0.82 & 0.84 & 0.87 & 0.72 & 0.42 & 0.73 & 1.01 & 0.73 & 1.00 & 1.13 & 0.63 & 0.71 \\
nA2 & 0.92 & 0.91 & \textbf{0.98} & 0.77 & 0.37 & \textbf{0.79} & 1.14 & 0.81 & \textbf{1.14} & 1.28 & 0.64 & \textbf{0.75} \\
%nAvg (A) & 1.18 & 0.84 & 1.21 & 0.95 & 0.21 & 0.95 & 1.34 & 0.76 & 1.34 & 1.63 & 0.71 & 0.86 \\
\midrule
\multicolumn{13}{>{\hsize=10\hsize}X}{\textbf{English participants}} \\
\midrule
R9 (B) & 3.42 & 3.88 & 3.60 & 3.11 & 6.50 & 3.45 & 2.75 & 3.25 & 2.50 & 5.50 & 1.75 & 3.15 \\ % B3
R10 (B) & 2.80 & 2.90 & 2.85 & 3.15 & 1.43 & 2.55 & 4.60 & 2.13 & 2.75 & 1.375 & 1.67 & 1.55 \\ % QX7
%--
R11 (I) & 1.71 & 3.00 & 2.10 & 1.76 & 2.33 & 1.85 & 5.29 & 2.46 & 3.45 & 4.50 & 1.89 & 2.15 \\ % A5
%--
R12 (A) & 2.80 & 4.80 & 3.30 & 3.19 & 3.75 & 3.30 & 5.40 & 2.90 & 4.15 & 2.00 & 1.74 & 1.75 \\ % xz2
nA & 0.60 & 0.81 & 0.66 & 0.63 & 0.73 & 0.62 & 1.04 & 0.59 & 0.73 & 0.73 & 0.40 & 0.47 \\
nA2 & 0.50 & 0.86 & \textbf{0.60} & 0.54 & 0.67 & \textbf{0.57} & 1.20 & 0.60 & \textbf{0.85} & 0.76 & 0.41 & \textbf{0.44} \\
\midrule
\multicolumn{13}{>{\hsize=10\hsize}X}{\textbf{Other participants}} \\
\midrule
R13 (I) & 2.44 & 1.75 & 2.30 & 1.50 & 3.00 & 1.65 & 1.44 & 1.82 & 1.65 & 3.80 & 1.93 & 2.40 \\ % aw8
%--
R14 (A) & 2.30 & 0 & 2.30 & 1.90 & 2.00 & 1.90 & 2.15 & 0 & 2.15 & 2.00 & 2.06 & 2.05 \\ % B3
R15 (A) & 1.21 & 2.00 & 1.25 & 1.55 & 0 & 1.55 & 1.58 & 2.00 & 1.60 & 3.00 & 1.22 & 1.40 \\ % 90
nA\footnote{Since this group does not contain any beginner learners, the normalized average corresponds to nAvg (I+A). However, due to the mixed nature of this group, these results should be taken with a grain of salt; we included the normalized average for completeness' sake.} & 0.81 & 0.53 & \textbf{0.80} & 0.70 & 0.64 & \textbf{0.72} & 0.74 & 0.54 & \textbf{0.76} & 1.22 & 0.72 & \textbf{0.80} \\
\bottomrule
\end{tabularx}
\end{minipage}
\caption{Results by rater and mother tongue}\label{tab:res2}
\end{table*}

Beginning learners seem not to show any predictable pattern. However, if we look at intermediate and advanced learners of Swedish, we see some interesting patterns.

For German speakers, we can see that on average (nAvg (I+A)) their rejection time of German-looking non-words (D$0$) is higher than their rejection time of English-looking non-words (E$0$). On average their reaction time to German-looking non-words (DC) is also higher than their reaction time to English-looking non-words (EC). 

For English speakers, rejection time of English-looking non-words (E$0$) is slightly higher than rejection time of German-looking non-words (D$0$).

For non-target participants, we can see no clear pattern. However, they never react fastest on existing Swedish words. The fastest reaction times are on English non-words for raters R13 and R14, possibly due to the larger typological distance between Swedish and English. Rater R15 was marginally faster on rejection of German non-words than on recognition of Swedish words.

German and English intermediate and advanced participants follow the expected pattern of recognizing existing Swedish words fastest and reacting slowest on Swedish looking non-words. However, this small scale study hints at the possibility that L1-specific non-words are processed second-slowest, possibly due to the activation of the native language system.
However, the low number of participants in this study does not allow us to draw any reliable conclusions.

%----

% Problematic words
% kanden => fil kand => abbreviation and wrong expansion of kand
% handes => wrong non existent declension of "hA"
% grande dame => not swedish

% => use Saldo lemmata as source for SV words instead of "more exhaustive" saldom

% -----

% User interface/API for querying
\section{User interface and API}
\begin{figure}
\includegraphics[width=\textwidth]{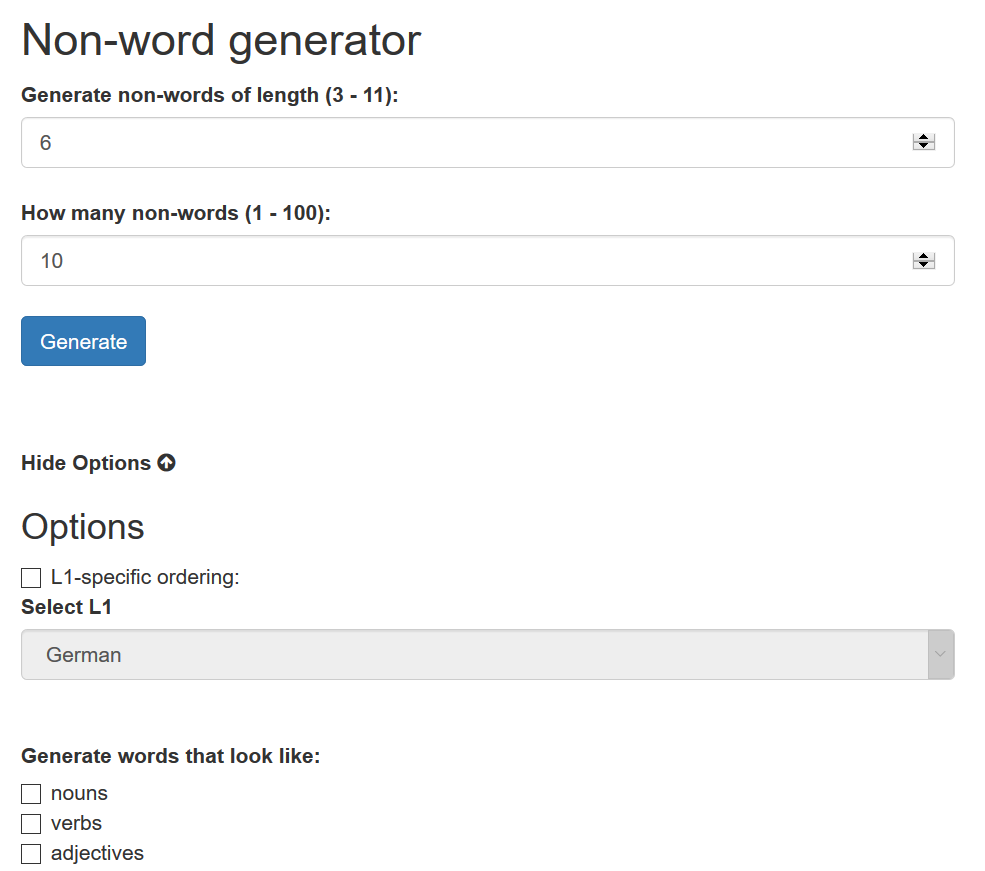}
\caption{Graphical user interface with some options visible}\label{fig:ui}
\end{figure}

We have created a web-based graphical user interface that facilitates generation of non-words as shown in figure \ref{fig:ui}.\footnote{\url{https://spraakbanken.gu.se/larkalabb/nonwords}} 
The interface is kept simple, with the standard view offering a choice of length of non-words and the number of non-words to generate, including sensible default values as well as boundaries.
It is also possible to access additional options such as L1-specific language ordering and the possibility to generate words that look like nouns, verbs, adjectives or adverbs.

We also provide a REST API (Application Programming Interface) for easy integration with other services. %Further details will be included after  review.
% Details provided after anonymous period

% -----
\section{Discussion}
During our experiments, we encountered the notoriously difficult-to-solve problem of accidentally generating existing words. These generated words might be colloquial, slang or non-standard words, technical jargon, words used only in certain areas, names, abbreviations, compounds, and many more. In order to exclude more existing words, we could either use a larger resource to check against or use the language model to find neither too highly rated nor too poorly rated non-words, in the hope that by excluding highly rated words, more existing words are excluded. However, this would probably not exclude rare words. Another approach would be to manually sift through all generated non-words and exclude all existing words. However, that would be labor-intensive; crowd-sourcing could potentially help with this task.

Even though accuracy in recognition of existing Swedish words was not the aim of this study, post-experimental interviews revealed that a lot of people had problems with words that could be confused for a misspelled version of an existing word despite the word in question being an existing word as well, such as for example the word \emph{komman}. If analyzed as the commonly used verb \emph{komma} `to come' with an \emph{n} at the end, rejection rates are high. However, it can also be analyzed as noun \emph{komma} `comma', of which \emph{komman} is the definite form (viz. \emph{the} comma). Some people also had difficulties with spelling, not being sure how exactly a word was spelled. Finally, there were some problematic words such as \emph{*kanden}, which, according to Saldom should be an existing Swedish word but was not identified as Swedish word by any of the native speakers. This is due to the automatic paradigm expansion used in Saldom which, for the entry \emph{fil. kand.} (= Filosofie kandidat) `Bachelor of Arts' generated `kanden' from the second part. Similarly, non-Swedish words and expressions present in Saldom, such as \emph{grande dame}, lead to nonsensical forms.

External validity certainly is a concern in these studies, as the number of participants as well as their selection was not randomized. However, results seem promising. Internal validity could also be problematized, as part of the instructions, for example in study 2, where participants where asked to perform a word recognition task, could have biased them towards more `actively' activating their L2 mental lexicon. However, as the participants were unaware that the variable being measured was their reaction time to L1-specific non-words, we claim that the internal validity is sufficiently high.

%\section{Future work}
% allow consonant clusters in onset
% SV allows up to three C in onset, of which the first is fricative \citep{}??
% -> use random first, bigram model for second, continue as before with trigram, then qgrams
%In the future, we would like to explore L1-specific non-words in a bigger context, including more matrix languages and more distractor languages.

%We also plan on integrating this work in a diagnostic language test and assess its validity with language learners of different L1 backgrounds.
\newpage

\bibliographystyle{apalike}
\bibliography{naaclhlt2018}
\end{document}